\documentclass{article}

\usepackage{microtype}
\usepackage{graphicx}
\usepackage{subcaption}
\usepackage{booktabs} 

\usepackage{hyperref}

\usepackage{natbib}
\usepackage[accepted]{icml2025/icml2025}

\usepackage{amsmath}
\usepackage{amssymb}
\usepackage{mathtools}
\usepackage{amsthm}
\usepackage{multirow}

\usepackage[capitalize,noabbrev]{cleveref}
\usepackage{enumitem}
\theoremstyle{plain}

\theoremstyle{definition}

\theoremstyle{remark}

\usepackage[textsize=tiny]{todonotes}

\usepackage{amsmath,amsfonts,bm}
\usepackage{mathtools}

\def\eqref#1{equation~\ref{#1}}

\def\1{\bm{1}}

\def\vz{{\bm{z}}}

\def\mA{{\bm{A}}}

\def\mI{{\bm{I}}}

\DeclareMathAlphabet{\mathsfit}{\encodingdefault}{\sfdefault}{m}{sl}
\SetMathAlphabet{\mathsfit}{bold}{\encodingdefault}{\sfdefault}{bx}{n}

\def\sR{{\mathbb{R}}}
\def\sS{{\mathbb{S}}}

\DeclareMathOperator*{\argmin}{arg\,min}

\makeatletter
\renewcommand{\paragraph}{%
  \@startsection{paragraph}{4}%
  {\z@}{0.25ex \@plus 0.25ex \@minus .5ex}{-1em}%
  {\normalfont\normalsize\bfseries}%
}
\makeatother

\DeclareMathOperator{\WER}{WER}
\DeclareMathOperator{\RTF}{RTF}

\icmltitlerunning{Whisper with Sparse Encoder}

\begin{document}

\twocolumn[
\icmltitle{Early Attentive Sparsification Accelerates Neural Speech Transcription}



\icmlsetsymbol{equal}{*}

\begin{icmlauthorlist}
\icmlauthor{Zifei Xu}{dmx}
\icmlauthor{Sayeh Sharify}{dmx}
\icmlauthor{Hesham Mostafa}{dmx}
\icmlauthor{Tristan Webb}{dmx}
\icmlauthor{Wanzin Yazar}{dmx}
\icmlauthor{Xin Wang}{dmx}

\end{icmlauthorlist}

\icmlaffiliation{dmx}{d-Matrix, Santa Clara, USA}

\icmlcorrespondingauthor{Zifei Xu}{xuzifei@d-matrix.ai}
\icmlcorrespondingauthor{Xin Wang}{xwang@d-matrix.ai}

\icmlkeywords{Machine Learning, ICML}

\vskip 0.3in
]



\printAffiliationsAndNotice{}  

\begin{abstract}

    Transformer-based neural speech processing has achieved state-of-the-art performance.  
    Since speech audio signals are known to be highly compressible, here we seek to accelerate neural speech transcription by time-domain signal sparsification early in the neural encoding stage, taking advantage of the interpretability of the self-attention mechanism in transformer audio encoders. 
    With the Whisper family of models, we perform a systematic architecture search over the joint space of sparsification stage (a certain encoder layer) and compression ratio (sparsity).
    We found that the best resulting solutions under $1\%$ accuracy degradation choose to sparsify the hidden state to $40\text{-}60\%$ sparsity at an early encoding stage, and thereby achieve up to $1.6\times$ runtime acceleration in English speech transcription tasks on Nvidia GPUs without any fine-tuning.  

\end{abstract}
\section{Introduction}
Transformer-based automatic speech recognition (ASR) systems have set recent state-of-the-art records on numerous speech transcription benchmarks~\cite{whisper}. 
Unlike older approaches that required special audio-text alignment during training~\cite{CTC,deepspeech}, or that trained the audio model independently from the language model~\cite{wav2vec}, transformer architectures use an encoder-decoder architecture to encode the audio sequence and then decode the text transcription from the audio sequence embeddings. The self-attention and cross-attention mechanisms in the encoder and decoder allow information to flow between audio tokens, and from audio tokens to text tokens. The Whisper model family is a popular and highly-performant transformer-based ASR model family. In this paper, we investigate token-dropping strategies to reduce the number of tokens in the encoder stack of Whisper models.


Token-eviction or token-dropping methods have traditionally been applied to reduce the size of the KV (key-value) cache in large language models~\cite{H2O,Qhitter}. They are driven by the observation that language token sequences are often highly redundant with several tokens that contribute little information to the language modeling task. One common approach to identify these unimportant tokens is to use their cumulative attention scores across heads and source tokens~\cite{Spatten,H2O}. These cumulative scores indicate how strongly these tokens are used (aggregated) by other tokens. However, it is still an open question whether such token redundancy is present in tokenized audio signals, and if such redundancy exists, what the optimal token dropping strategies are that can exploit this redundancy to reduce the compute and memory cost of the audio encoder.  

In this paper, our goal is to exploit possible audio token redundancies in the Whisper model family. We analyze the attention scores across multiple layers to obtain insights into how the importance of different audio tokens varies across layers. We experiment with several token dropping strategies. We construct a Pareto front of token dropping recipes that achieve optimal tradeoffs between WER (Word Error Rate) and inference latency. We show that these recipes generalize well across multiple Whisper model variants.

\begin{figure*}[t!]
\vskip 0.2in
\begin{center}
\centerline{\includegraphics[width=4.0in]{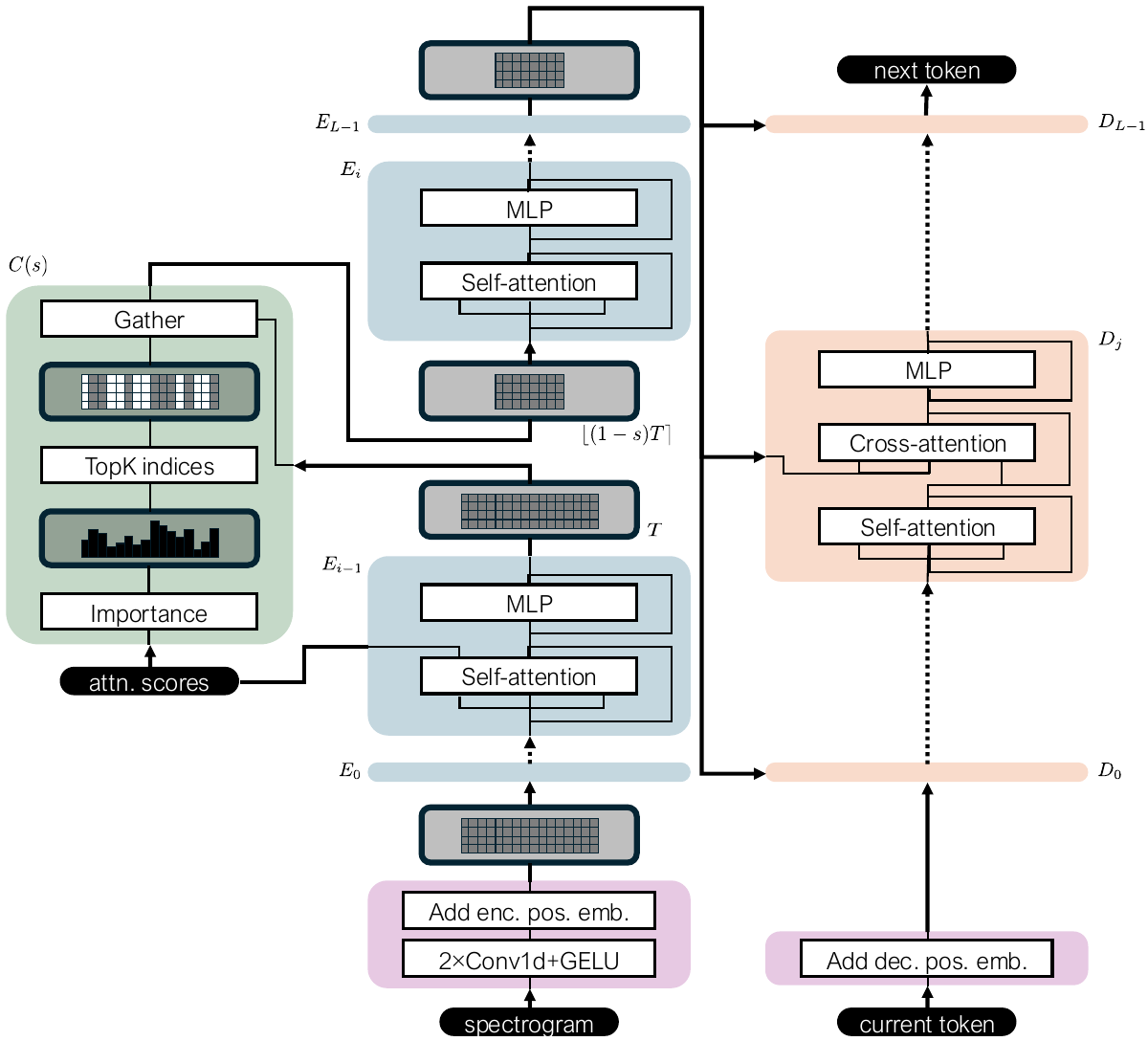}}
\caption{
    Architecture of the Whisper model modified with early attentive sparsification.  
    The original Whisper model is composed of an encoder stack $E$'s (middle column) and a decoder stack $D$'s (right column), of depth $L$.  
    Flowing through the encoder stack are hidden states of sequence dimension $T$, determined by embedded audio signal duration. 
    At an early encoding stage $i \in \{1,\cdots,L\}$, we compress the hidden state by a time-domain sparsifier $C$ (left column, see Section~\ref{sec:methods} for details) at sparsity $s\in[0,1)$, to a reduced sequence dimension $\lfloor(1-s)T\rceil$, which persists in the rest of the encoder stack and in the subsequent decoder stack.
    In the choice of $(i,s)$ consists the architectural search space.  
}
\label{fig:arch}
\end{center}
\vskip -0.2in
\end{figure*}

\section{Related work}

\subsection{Encoder-decoder transformer models}

The transformer~\citep{vaswani2017attention} architecture's ability to model long range dependencies has made it the method of choice in many domains with sequential input. Audio data, in particular, can be naturally represented as a temporal sequence of audio samples
Early models such as Wave2Vec~\citep{wav2vec} and HuBERT~\citep{hsu2021hubert} extended the Transformer architecture and utilized large audio datasets~\citep{librispeech}. These models trained directly end-to-end on raw audio input spectrograms in a self-supervised manner. 
\citealt{whisper} took advantage of the underappreciated nature of weakly supervised pre-training, leading to Whisper models, a staged multi-task training pipeline which processed 680,000 hours of multi-lingual audio data and paired text transcripts during pre-training. 

The encoder-decoder transformer block architecture~(consisting of attention, feedforward, and convolutional layers) has formed the backbone for numerous other ASR models. 
Architectures used in recent industrial applications~\citep{zhang2023google, ramirez2024anatomy} consist of encoder-decoder architecture comprised of conformer blocks~\citep{gulati2020conformer, rekesh2023fast} in the encoder, and RNN-T layers~\citep{graves2012sequence, ghodsi2020rnn, kuang2022pruned} in the decoder stack. 
The encoder stack is more critical to runtime efficiency than the decoder stack~\citep{prabhavalkar2023end}. 
Others have proposed elimination of convolutions entirely from the architecture, instead relying solely on self-attention~\citep{hou2024revisiting}.

\section{Methods}
\label{sec:methods}

\subsection{Models and benchmarking tasks}
We experiment with 10 publicly available models from the HuggingFace Hub with parameter counts spanning three orders of magnitude: the OpenAI Whisper family~\cite{whisper} and the HuggingFace Distil-Whisper family~\cite{Distil-whisper}.  
These models are \texttt{whisper-tiny} (37.8M), \texttt{whisper-base} (72.6M), \texttt{whisper-small} (242M), \texttt{whisper-medium} (764M), \texttt{whisper-large-v3-turbo} (809M), \texttt{whisper-large-v3} (1.54B), \texttt{distil-small.en} (166M), \texttt{distil-medium.en} (394M), \texttt{distil-large-v3.5} (756M) and \texttt{distil-large-v3} (756M). \\

The models are benchmarked on the Librispeech ASR automatic speech transcription task~\cite{librispeech} yielding word error rate (WER) as a performance metric. 
A subset, 300 task examples, of the clean validation split is used for architecture search
(see Appendix~\ref{app:datasize} for analysis on the adequacy of the choice of this data set size).

\subsection{Early attentive sparsification (EAS) mechanism}
As shown in Figure~\ref{fig:arch}, we sparsify the hidden state $\vz^l\in\sR^{N \times T}$ at an early encoder stage $l=i\in\{1,\cdots,L\}$ to a sparsity level $s\in[0,1)$, where $N$ is the embedding dimension and $T$ the number of audio samples in the time domain. 
The sparsification mechanism is based on top-k selection of a heuristic \emph{importance score} $\mI^l \in \sR ^{T}$ computed from the previous encoder layer's (post-softmax) self-attention score $\mA^{l-1} \in \sR^{H \times T \times T}$ ($H$ denoting the number of attention heads) as the arithmetic mean across attention heads 
\begin{align}
    I^l_t &= \frac {1} {HT} \sum_{h,t'} {A^{l-1}_{h,t',t}} . 
\end{align} 
The sparsifier effectively reduces the temporal sequence length from $T$ to $\lfloor (1-s)T \rceil$ at encoding stage $l=i$,     
\begin{align}
    \vz^i &\leftarrow \mathtt{gather}\left(\vz^i, \underset{t|k=\lfloor (1-s)T \rceil}{\mathtt{argtopk}} I^i_t\right) . 
\end{align} 


\subsection{Runtime speedup measurement}
The aforementioned sparsification mechanism impacts the inference-time computational efficiency in two ways: applying the sparsifier introduces a constant overhead at an encoder stage, and from then on the reduced hidden state temporal dimension speeds up the rest of the model execution. 
As such, the earlier the sparsification stage $i$ is, and the higher the sparsity $s$ is, the faster the overall computation will be.  

To assess the practical runtime speedup resulting from architecture choice $(i,s)$, we measure wall-clock time of model inference on a single Nvidia A100 GPU with 80GB HBM. 
We record the encoding and decoding time for individual instances of the transcription task, and quantify \emph{real time factor} (RTF, \citealt{open-asr-leaderboard}), defined as the dimensionless ratio between the total inference time and the total duration of the raw speech audio signals. 
The lower the $\RTF$, the faster the transcription.  
Relative speedup is calculated as $\frac{\RTF_0}{\RTF}$, $\RTF_0$ being the baseline RTF.  
Note that, though other independent optimizations such as using lower numerical precision, data batching, low-rank weight compression~\citep{liteasr}, and FlashAttention-2~\cite{dao2023flashattention2fasterattentionbetter} could push RTF to even lower levels, in addition to our technique, the relative speedup due to early attentive sparsification persists in the presence of other orthogonal runtime optimizations (see Appendix~\ref{app:other_optimization}).  


\subsection{Architecture search}
\label{sec:ArchitecturalSearch}
We perform a grid search of sparsification stage and sparsity $(i, s) \in [L] \times \sS$, where $[L] \triangleq \{1,...L\}$, and $\sS=\{0.0, 0.1, 0.2, 0.3, 0.4, 0.5, 0.6, 0.7, 0.8, 0.9\}$. 
Along the empirical Pareto front, we minimize the overall real time factor $\RTF$ constrained by accuracy, \emph{viz.} $1-\WER$, degrades no more than $1\%$ relative to the baseline.
Formally, the optimization problem is
\begin{equation}
\begin{aligned}
(i^*,s^*) &= \underset{(i, s) \in [L] \times \sS}{\argmin} \;\RTF(i,s) ,
\\
&\text{s.t.}\;
\{ (i', s') | \WER(i',s') < \WER(i,s), \\
& \RTF(i',s')< \RTF(i,s) \} = \varnothing,
\\
&\text{and}\;
1 - \WER(i,s)
\;\ge\;
0.99\,\left(1 - \WER_{0}\right), 
\end{aligned}
\label{eq:constraint}
\end{equation}
where $\WER_{0}$ is the word error rate achieved by the baseline model.

\begin{figure*}[h!]
\vskip 0.2in
\begin{center}
\centerline{\includegraphics[width=6in]{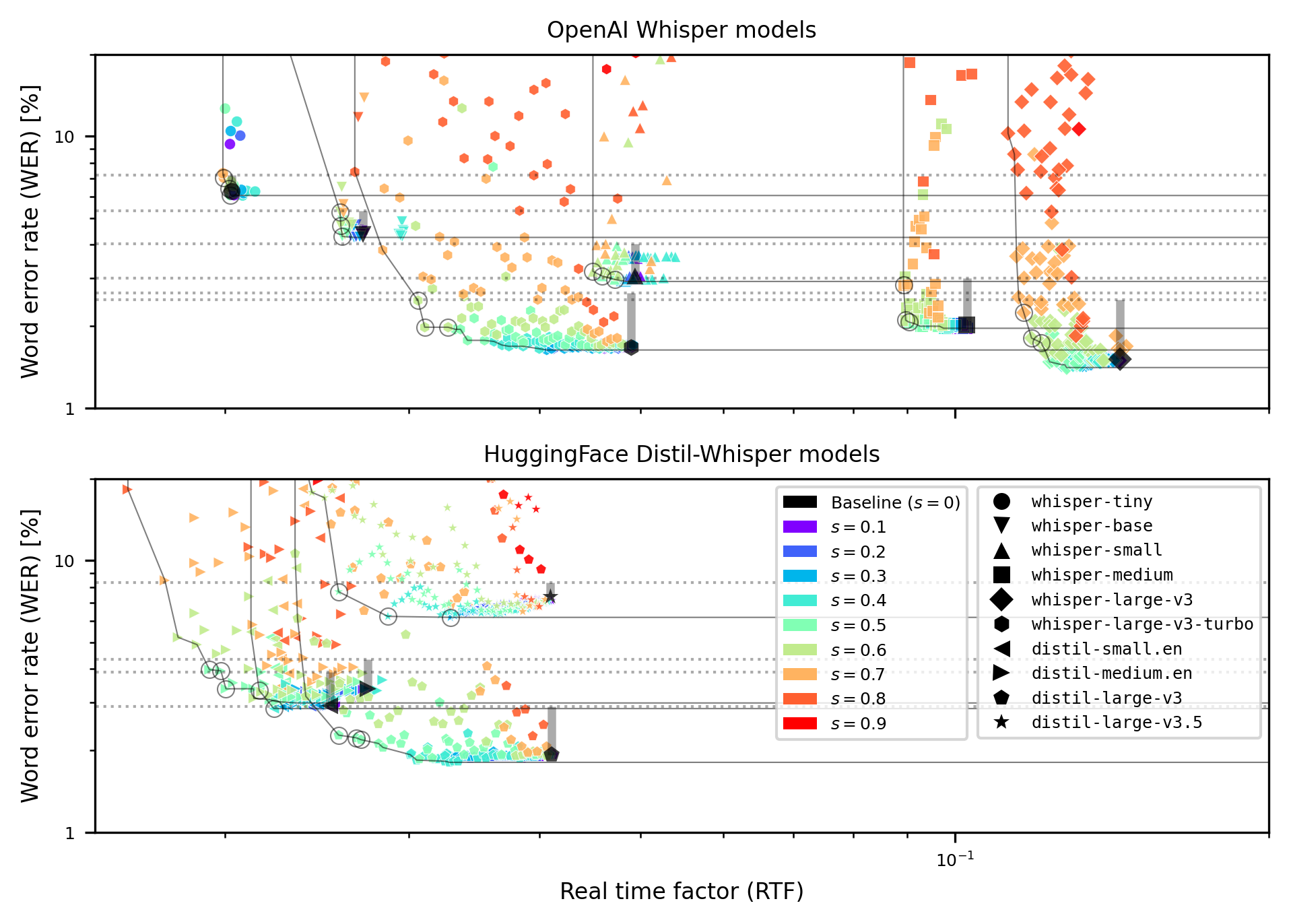}}
\caption{
    Empirical results of architecture search for OpenAI's Whisper (top) and HuggingFace's Distil-Whisper (bottom) models, showing tradeoff between WER and RTF in the context of the benchmark transcription task. 
    Each colored symbol represents a unique $(i, s)$ configuration (see Section~\ref{sec:methods}) for a certain model (identity of the model encoded by symbol shape).
    Large, black symbols represent the baseline configuration for each model, from which a thick, vertical bar rises to 1\% accuracy degradation (Equation~\ref{eq:constraint}) from baseline, leading to the 99\% accuracy boundary (horizontal dotted lines) serving as a constraint of the optimization. 
    Empirical Pareto fronts of each model are marked by thin, gray curves. 
    We identify top-3 configurations (circled) on the Pareto fronts of each model that satisfy the accuracy constraint as admissible solutions to report in Table~\ref{table:WER}.
}
    \label{fig:opt_res}
\end{center}
\vskip -0.2in
\end{figure*}

\begin{table}[h!]  
\centering
\scriptsize
\begin{tabular}{l|l|l|l}
    \toprule
        Model & $(i,s)$ & WER [\%] & RTF \\
        \hline \hline
       \multirow{3}{*}{\tiny{\texttt{whisper-tiny}}} & Baseline & 6.266 &0.020 \\
        & (3, 0.7) & 7.050 (0.992) & 0.020 (1.018$\times$)\\
        \multirow{1}{*}{\tiny{(37.8M, 4)}}& (2, 0.6) & 6.406 (0.999) & 0.020 (1.007$\times$)\\
        & (2, 0.5) & 6.075 (1.002) & 0.020 (1.003$\times$)\\
        \hline
        \multirow{3}{*}{\tiny{\texttt{whisper-base}}} & Baseline & 4.352 &0.027 \\
        & (1, 0.6) & 5.274 (0.990) & 0.026 (1.052$\times$)\\
        \multirow{1}{*}{\tiny{(72.6M, 6)}}& (3, 0.6) & 4.682 (0.997) & 0.026 (1.050$\times$)\\
        & (4, 0.5) & 4.282 (1.001) & 0.026 (1.047$\times$)\\
        \hline
        \multirow{3}{*}{\tiny{\texttt{whisper-small}}} & Baseline & 3.064 &0.049 \\
        & (1, 0.6) & 3.185 (0.999) & 0.045 (1.097$\times$)\\
        \multirow{1}{*}{\tiny{(242M, 12)}}& (1, 0.5) & 3.064 (1.000) & 0.046 (1.076$\times$)\\
        & (6, 0.5) & 2.977 (1.001) & 0.047 (1.046$\times$)\\
        \hline
        \multirow{3}{*}{\tiny{\texttt{whisper-medium}}} & Baseline & 2.019 &0.103 \\
        & (7, 0.7) & 2.837 (0.992) & 0.089 (1.150$\times$)\\
        \multirow{1}{*}{\tiny{(764M, 24)}}& (1, 0.6) & 2.106 (0.999) & 0.090 (1.145$\times$)\\
        & (1, 0.5) & 2.071 (0.999) & 0.090 (1.136$\times$)\\
        \hline
        \multirow{3}{*}{\tiny{\texttt{whisper-large-v3-turbo}}} & Baseline & 1.671 &0.049 \\
        & (2, 0.6) & 2.489 (0.992) & 0.031 (1.601$\times$)\\
        \multirow{1}{*}{\tiny{(809M, 32)}}& (3, 0.6) & 1.984 (0.997) & 0.031 (1.577$\times$)\\
        & (6, 0.6) & 1.984 (0.997) & 0.033 (1.498$\times$)\\
        \hline
        \multirow{3}{*}{\tiny{\texttt{whisper-large-v3}}} & Baseline & 1.514 &0.144 \\
        & (3, 0.7) & 2.245 (0.993) & 0.116 (1.237$\times$)\\
        \multirow{1}{*}{\tiny{(1.54B, 32)}}& (1, 0.6) & 1.810 (0.997) & 0.118 (1.216$\times$)\\
        & (3, 0.6) & 1.741 (0.998) & 0.121 (1.190$\times$)\\
        \hline
        \multirow{3}{*}{\tiny{\texttt{distil-small.en}}} & Baseline & 2.924 &0.025 \\
        & (1, 0.5) & 3.325 (0.996) & 0.022 (1.168$\times$)\\
        \multirow{1}{*}{\tiny{(166M, 12)}}& (1, 0.4) & 2.855 (1.001) & 0.022 (1.131$\times$)\\
        \hline
        \multirow{3}{*}{\tiny{\texttt{distil-medium.en}}} & Baseline & 3.377 &0.027 \\
        & (1, 0.5) & 3.969 (0.994) & 0.019 (1.418$\times$)\\
        \multirow{1}{*}{\tiny{(394M, 24)}}& (2, 0.5) & 3.934 (0.994) & 0.020 (1.383$\times$)\\
        & (3, 0.5) & 3.377 (1.000) & 0.020 (1.368$\times$)\\
        \hline
        \multirow{3}{*}{\tiny{\texttt{distil-large-v3}}} & Baseline & 1.932 &0.041 \\
        & (1, 0.5) & 2.280 (0.996) & 0.026 (1.600$\times$)\\
        \multirow{1}{*}{\tiny{(756M, 32)}}& (3, 0.5) & 2.228 (0.997) & 0.027 (1.538$\times$)\\
        & (4, 0.5) & 2.193 (0.997) & 0.027 (1.523$\times$)\\
        \hline
        \multirow{3}{*}{\tiny{\texttt{distil-large-v3.5}}} & Baseline & 7.371 &0.041 \\
        & (1, 0.5) & 7.686 (0.997) & 0.026 (1.596$\times$)\\
        \multirow{1}{*}{\tiny{(756M, 32)}}& (1, 0.4) & 6.236 (1.012) & 0.029 (1.431$\times$)\\
        & (2, 0.3) & 6.183 (1.013) & 0.033 (1.247$\times$)\\

    \bottomrule
        
\end{tabular}
\caption{
WER and RTF of the top-3~\protect\footnotemark{}
 architecture search results for each model. 
Parenthesized numbers show relative changes from baseline.  Under each model name we marked the parameter count of the model and number of encoder layers.
}
\label{table:WER}
\end{table}

\section{Experimental results}
\label{sec:res}

We present empirical optimization results for the OpenAI Whisper and the HuggingFace Distil-Whisper models in Figure~\ref{fig:opt_res}, and report top solutions in Table~\ref{table:WER}. 

The top solutions from the architecture search have sparsities between 0.4 and 0.6 at a very early (typically among the first 3) encoding stage, indicating a high level of task-irrelevant redundancy in the embedded hidden state.  
This is particularly amenable to runtime efficiency, resulting in up to 1.6$\times$ speedup from baseline. 


Notably, the highest speedup is achieved in Whisper models produced by post-training compression, suggesting a synergy between these compression techniques and our early attentive sparsification. 
First, \texttt{whisper-large-v3-turbo}, a further fine-tuned \texttt{whisper-large-v3} with a much lighter decoder, EAS leads to 1.60$\times$ speedup when it is only 1.24$\times$ for \texttt{whisper-large-v3}.  
Second, EAS results in even higher speedups in Distil-Whisper models than in their larger, teacher Whisper models; for example, EAS achieved 1.60$\times$ speedup in \texttt{distil-large-v3} as compared to 1.24$\times$ for \texttt{whisper-large-v3}, its teacher.  
These findings suggest that, EAS might be complementary or even synergistic to existing efficiency optimization techniques in speeding up neural ASR.


\footnotetext{ There are only 2 configurations in the search grid that satisfy the optimization constraints for \texttt{distil-small.en}.}
\section{Conclusion}

We introduce early attentive sparsification (EAS), a simple mechanism that shortens the hidden state in Whisper-based ASR models early in the encoding stage based on local self-attention scores.  
Without any fine-tuning, our method prunes 40-60\% of audio samples at an early encoder layer while maintaining $\geq 99\%$ of the baseline accuracy in speech transcription tasks. 
This results in up to $1.6\times$ speedups on a single A100 GPU. 
Our method is complementary to other efficiency optimization techniques, holding the promise in further improving inference-time efficiency of neural ASR.  

\newpage
\bibliography{references}

\begin{thebibliography}{23}
\providecommand{\natexlab}[1]{#1}
\providecommand{\url}[1]{\texttt{#1}}
\expandafter\ifx\csname urlstyle\endcsname\relax
  \providecommand{\doi}[1]{doi: #1}\else
  \providecommand{\doi}{doi: \begingroup \urlstyle{rm}\Url}\fi

\bibitem[Radford et~al.(2023)Radford, Kim, Xu, Brockman, McLeavey, and Sutskever]{whisper}
Alec Radford, Jong~Wook Kim, Tao Xu, Greg Brockman, Christine McLeavey, and Ilya Sutskever.
\newblock Robust speech recognition via large-scale weak supervision.
\newblock In \emph{International conference on machine learning}, pages 28492--28518. PMLR, 2023.

\bibitem[Graves et~al.(2006)Graves, Fern{\'a}ndez, Gomez, and Schmidhuber]{CTC}
Alex Graves, Santiago Fern{\'a}ndez, Faustino Gomez, and J{\"u}rgen Schmidhuber.
\newblock Connectionist temporal classification: labelling unsegmented sequence data with recurrent neural networks.
\newblock In \emph{Proceedings of the 23rd international conference on Machine learning}, pages 369--376, 2006.

\bibitem[Amodei et~al.(2016)Amodei, Ananthanarayanan, Anubhai, Bai, Battenberg, Case, Casper, Catanzaro, Cheng, Chen, et~al.]{deepspeech}
Dario Amodei, Sundaram Ananthanarayanan, Rishita Anubhai, Jingliang Bai, Eric Battenberg, Carl Case, Jared Casper, Bryan Catanzaro, Qiang Cheng, Guoliang Chen, et~al.
\newblock Deep speech 2: End-to-end speech recognition in english and mandarin.
\newblock In \emph{International conference on machine learning}, pages 173--182. PMLR, 2016.

\bibitem[Baevski et~al.(2020)Baevski, Zhou, Mohamed, and Auli]{wav2vec}
Alexei Baevski, Yuhao Zhou, Abdelrahman Mohamed, and Michael Auli.
\newblock wav2vec 2.0: A framework for self-supervised learning of speech representations.
\newblock \emph{Advances in neural information processing systems}, 33:\penalty0 12449--12460, 2020.

\bibitem[Zhang et~al.(2023{\natexlab{a}})Zhang, Sheng, Zhou, Chen, Zheng, Cai, Song, Tian, R{\'e}, Barrett, et~al.]{H2O}
Zhenyu Zhang, Ying Sheng, Tianyi Zhou, Tianlong Chen, Lianmin Zheng, Ruisi Cai, Zhao Song, Yuandong Tian, Christopher R{\'e}, Clark Barrett, et~al.
\newblock H2o: Heavy-hitter oracle for efficient generative inference of large language models.
\newblock \emph{Advances in Neural Information Processing Systems}, 36:\penalty0 34661--34710, 2023{\natexlab{a}}.

\bibitem[Zhang et~al.(2024)Zhang, Liu, Chen, Kailkhura, Chen, and Wang]{Qhitter}
Zhenyu Zhang, Shiwei Liu, Runjin Chen, Bhavya Kailkhura, Beidi Chen, and Atlas Wang.
\newblock Q-hitter: A better token oracle for efficient llm inference via sparse-quantized kv cache.
\newblock \emph{Proceedings of Machine Learning and Systems}, 6:\penalty0 381--394, 2024.

\bibitem[Wang et~al.(2021)Wang, Zhang, and Han]{Spatten}
Hanrui Wang, Zhekai Zhang, and Song Han.
\newblock Spatten: Efficient sparse attention architecture with cascade token and head pruning.
\newblock In \emph{2021 IEEE International Symposium on High-Performance Computer Architecture (HPCA)}, pages 97--110. IEEE, 2021.

\bibitem[Vaswani et~al.(2017)Vaswani, Shazeer, Parmar, Uszkoreit, Jones, Gomez, Kaiser, and Polosukhin]{vaswani2017attention}
Ashish Vaswani, Noam Shazeer, Niki Parmar, Jakob Uszkoreit, Llion Jones, Aidan~N Gomez, {\L}ukasz Kaiser, and Illia Polosukhin.
\newblock Attention is all you need.
\newblock \emph{Advances in neural information processing systems}, 30, 2017.

\bibitem[Hsu et~al.(2021)Hsu, Bolte, Tsai, Lakhotia, Salakhutdinov, and Mohamed]{hsu2021hubert}
Wei-Ning Hsu, Benjamin Bolte, Yao-Hung~Hubert Tsai, Kushal Lakhotia, Ruslan Salakhutdinov, and Abdelrahman Mohamed.
\newblock Hubert: Self-supervised speech representation learning by masked prediction of hidden units.
\newblock \emph{IEEE/ACM transactions on audio, speech, and language processing}, 29:\penalty0 3451--3460, 2021.

\bibitem[Panayotov et~al.(2015)Panayotov, Chen, Povey, and Khudanpur]{librispeech}
Vassil Panayotov, Guoguo Chen, Daniel Povey, and Sanjeev Khudanpur.
\newblock Librispeech: An asr corpus based on public domain audio books.
\newblock In \emph{2015 IEEE International Conference on Acoustics, Speech and Signal Processing (ICASSP)}, pages 5206--5210, 2015.
\newblock \doi{10.1109/ICASSP.2015.7178964}.

\bibitem[Zhang et~al.(2023{\natexlab{b}})Zhang, Han, Qin, Wang, Bapna, Chen, Chen, Li, Axelrod, Wang, et~al.]{zhang2023google}
Yu~Zhang, Wei Han, James Qin, Yongqiang Wang, Ankur Bapna, Zhehuai Chen, Nanxin Chen, Bo~Li, Vera Axelrod, Gary Wang, et~al.
\newblock Google {USM}: Scaling automatic speech recognition beyond 100 languages.
\newblock \emph{arXiv preprint arXiv:2303.01037}, 2023{\natexlab{b}}.

\bibitem[Ramirez et~al.(2024)Ramirez, Chkhetiani, Ehrenberg, McHardy, Botros, Khare, Vanzo, Peyash, Oexle, Liang, et~al.]{ramirez2024anatomy}
Francis~McCann Ramirez, Luka Chkhetiani, Andrew Ehrenberg, Robert McHardy, Rami Botros, Yash Khare, Andrea Vanzo, Taufiquzzaman Peyash, Gabriel Oexle, Michael Liang, et~al.
\newblock Anatomy of industrial scale multilingual {ASR}.
\newblock \emph{arXiv preprint arXiv:2404.09841}, 2024.

\bibitem[Gulati et~al.(2020)Gulati, Qin, Chiu, Parmar, Zhang, Yu, Han, Wang, Zhang, Wu, et~al.]{gulati2020conformer}
Anmol Gulati, James Qin, Chung-Cheng Chiu, Niki Parmar, Yu~Zhang, Jiahui Yu, Wei Han, Shibo Wang, Zhengdong Zhang, Yonghui Wu, et~al.
\newblock Conformer: Convolution-augmented transformer for speech recognition.
\newblock \emph{arXiv preprint arXiv:2005.08100}, 2020.

\bibitem[Rekesh et~al.(2023)Rekesh, Koluguri, Kriman, Majumdar, Noroozi, Huang, Hrinchuk, Puvvada, Kumar, Balam, et~al.]{rekesh2023fast}
Dima Rekesh, Nithin~Rao Koluguri, Samuel Kriman, Somshubra Majumdar, Vahid Noroozi, He~Huang, Oleksii Hrinchuk, Krishna Puvvada, Ankur Kumar, Jagadeesh Balam, et~al.
\newblock Fast conformer with linearly scalable attention for efficient speech recognition.
\newblock In \emph{2023 IEEE Automatic Speech Recognition and Understanding Workshop (ASRU)}, pages 1--8. IEEE, 2023.

\bibitem[Graves(2012)]{graves2012sequence}
Alex Graves.
\newblock Sequence transduction with recurrent neural networks.
\newblock \emph{arXiv preprint arXiv:1211.3711}, 2012.

\bibitem[Ghodsi et~al.(2020)Ghodsi, Liu, Apfel, Cabrera, and Weinstein]{ghodsi2020rnn}
Mohammadreza Ghodsi, Xiaofeng Liu, James Apfel, Rodrigo Cabrera, and Eugene Weinstein.
\newblock Rnn-transducer with stateless prediction network.
\newblock In \emph{ICASSP 2020-2020 IEEE International Conference on Acoustics, Speech and Signal Processing (ICASSP)}, pages 7049--7053. IEEE, 2020.

\bibitem[Kuang et~al.(2022)Kuang, Guo, Kang, Lin, Luo, Yao, and Povey]{kuang2022pruned}
Fangjun Kuang, Liyong Guo, Wei Kang, Long Lin, Mingshuang Luo, Zengwei Yao, and Daniel Povey.
\newblock Pruned rnn-t for fast, memory-efficient asr training.
\newblock \emph{arXiv preprint arXiv:2206.13236}, 2022.

\bibitem[Prabhavalkar et~al.(2023)Prabhavalkar, Hori, Sainath, Schl{\"u}ter, and Watanabe]{prabhavalkar2023end}
Rohit Prabhavalkar, Takaaki Hori, Tara~N Sainath, Ralf Schl{\"u}ter, and Shinji Watanabe.
\newblock End-to-end speech recognition: A survey.
\newblock \emph{IEEE/ACM Transactions on Audio, Speech, and Language Processing}, 32:\penalty0 325--351, 2023.

\bibitem[Hou et~al.(2024)Hou, Huybrechts, Bhatia, Garcia-Romero, Han, and Kirchhoff]{hou2024revisiting}
Zejiang Hou, Goeric Huybrechts, Anshu Bhatia, Daniel Garcia-Romero, Kyu Han, and Katrin Kirchhoff.
\newblock Revisiting convolution-free transformer for speech recognition.
\newblock 2024.

\bibitem[Gandhi et~al.(2023)Gandhi, von Platen, and Rush]{Distil-whisper}
Sanchit Gandhi, Patrick von Platen, and Alexander~M Rush.
\newblock Distil-whisper: Robust knowledge distillation via large-scale pseudo labelling.
\newblock \emph{arXiv preprint arXiv:2311.00430}, 2023.

\bibitem[Srivastav et~al.(2023)Srivastav, Majumdar, Koluguri, Moumen, Gandhi, et~al.]{open-asr-leaderboard}
Vaibhav Srivastav, Somshubra Majumdar, Nithin Koluguri, Adel Moumen, Sanchit Gandhi, et~al.
\newblock Open automatic speech recognition leaderboard.
\newblock \url{https://huggingface.co/spaces/hf-audio/open_asr_leaderboard}, 2023.

\bibitem[Kamahori et~al.(2025)Kamahori, Kasai, Kojima, and Kasikci]{liteasr}
Keisuke Kamahori, Jungo Kasai, Noriyuki Kojima, and Baris Kasikci.
\newblock Liteasr: Efficient automatic speech recognition with low-rank approximation, 2025.
\newblock URL \url{https://arxiv.org/abs/2502.20583}.

\bibitem[Dao(2023)]{dao2023flashattention2fasterattentionbetter}
Tri Dao.
\newblock Flashattention-2: Faster attention with better parallelism and work partitioning, 2023.
\newblock URL \url{https://arxiv.org/abs/2307.08691}.

\end{thebibliography}

\newpage
\section{Appendix}

\subsection{Choice of data set size for the architectural search}
\label{app:datasize}
\begin{figure}[ht!]
\vskip 0.2in
\begin{center}
\centerline{\includegraphics[width=3in]{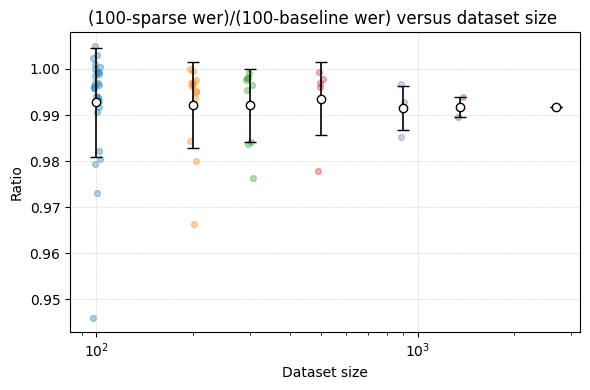}}
\caption{ Accuracy ratio between model with encoder sparsification and baseline model at different dataset sizes, where $accuracy=1-WER$. Hollow circles represent mean of accuracy degradation with error bars for one standard deviation.
}
\label{fig:dataset_size}
\end{center}
\end{figure}

The LibriSpeech ASR dataset \cite{librispeech} validation split has 2,703 audio examples. Due to the large search space across encoder layers and sparsity for our experiments, it is inefficient and impractical to run the whole validation split. To determine the smallest subset that still yields stable accuracy estimates, we ran experiments on the whole split on a \texttt{whisper-large-v3-turbo} model with drop layer 21 and sparsity 0.6 ( a sparsification configuration that provides reasonable speedup without much accuracy degradation), and the non sparsified baseline model. We partitioned the data split into groups with different sizes and plot the accuracy degradation versus dataset size, as shown in Fig. \ref{fig:dataset_size}. We selected 300 as the optimal dataset size since it is a reasonable size while its error bar falls within our tolerance of 0.01 accuracy degradation for selecting the optimal sparsification configuration, striking a balance between statistical reliability and computational cost.

\subsection{Aggregation function for importance score}
\label{app:agg_func}

We asked if any statistics than than arithmetic mean serve as a better aggregation function in computing importance scores from attention scores.  

\begin{figure*}[h!]
\vskip 0.2in
\begin{center}
\centerline{\includegraphics[width=5in]{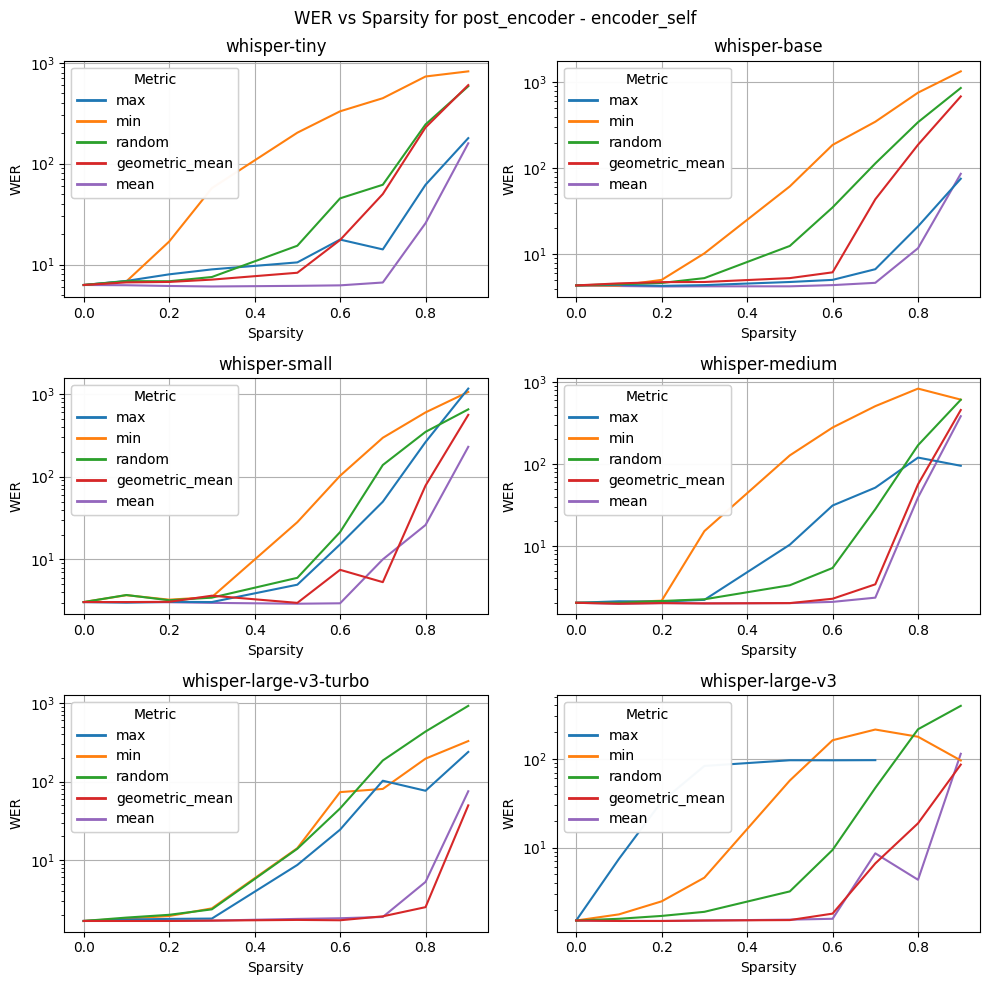}}
\caption{ WER versus sparsity for dropping at the last encoder layer using importance score aggregated across all encoder layers. Sub figures correspond to different models and colored lines represent various aggregation function, with $random$ meaning randomly dropping audio samples. 
}
\label{fig:agg_encoder}
\end{center}
\end{figure*}

To answer this question, we performed a search over the set for the choice of aggregation function. Fig.~\ref{fig:agg_encoder} and Table ~\ref{table:agg_func} shows the WER of models sparsified at the last encoder layer using importance scores aggregated across all encoder layers. Among the aggregation functions, $mean$ produces the lowest WER most of the time, hence it was selected to be used in our main experiments. For all model sizes, we observe that sparsity can go up to 0.6 without noticeable degradation in WER when using the $mean$ aggregation function.

\subsection{Runtime decomposition}
\begin{figure}[!htb]
\vskip 0.2in
\begin{center}
\centerline{\includegraphics[width=3in]{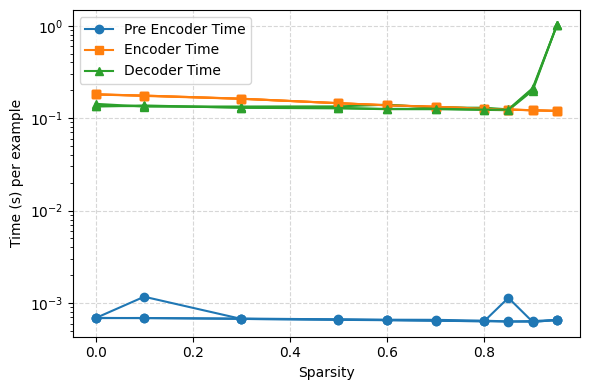}}
\caption{ Average time spent at different components of the Whisper model per example vs sparsity for \texttt{whisper-large-v3-turbo} model at sparsification layer of 20. To reduce the effect of system noise, the same experiment was repeated 3 times and plotted on top of each other. }
\label{fig:time}
\end{center}
\end{figure}

\begin{figure}[!htb]
\vskip 0.2in
\begin{center}
\centerline{\includegraphics[width=3in]{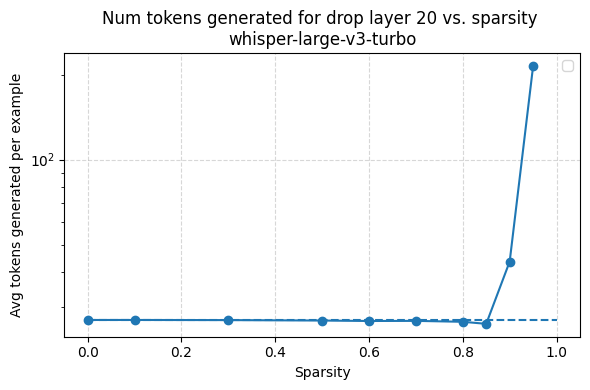}}
\caption{ Average number of tokens generated per example vs sparsity for \texttt{whisper-large-v3-turbo} model at sparsification layer of 20. The dashed line indicates the average number of tokens generated per example for the baseline model.
}
\label{fig:num_token}
\end{center}
\end{figure}

We profiled the time spent in each component of the Whisper model with increasing sparsity at a specific sparsification layer (layer 20) for \texttt{whisper-large-v3-turbo} in Fig. \ref{fig:time}. We find that time spent before the encoder stack (Conv and GELU ops) did not change with increasing sparsity while time spent in the encoder and decoder stack both decreases, at a slope where the former is steeper than the latter. There is a surge of decoder time at very high sparsities. This can be explained by the tendency to generate more text tokens when the model is at high sparsity, as depicted in Fig. \ref{fig:num_token}. For example, at sparsity=0.9, the model would generate \textit{"He also thought of his managerial position, his managerial position, his managerial position, his managerial position, his managerial position, his managerial position, his managerial position, his managerial position, his managerial position, his managerial position, his managerial position, his managerial position, his managerial position"}, instead of  \textit{"He also thought of his managerial position."} generated by the baseline model. This also agrees with our observation that when padded audios tokens are removed for shorter audio sequences, the model would have difficulty identifying the stopping point and generate indefinitely. 

\subsection{Compatibility with other efficient methods}
\label{app:other_optimization}
Our baseline real time factors (RTF) are higher than the figures reported on the OpenASR Leaderboard \cite{open-asr-leaderboard} because we benchmark with \texttt{float32} precision and a batch size of 1, whereas the leader board uses bf16 and large batches (64). Importantly, the proposed token-dropping scheme operates on sequence length, making it orthogonal to weight-space optimizations such as low-rank factorization \cite{liteasr}, pruning, or quantization, and complementary to kernel-level accelerators. Appendix \ref{app:attn} demonstrates that combining our method with FlashAttention-2 \cite{dao2023flashattention2fasterattentionbetter} or PyTorch SDPA yields additional speed-ups without compromising accuracy.

\begin{figure*}[tbh!]
\vskip 0.2in
\begin{center}
\centerline{\includegraphics[width=5in]{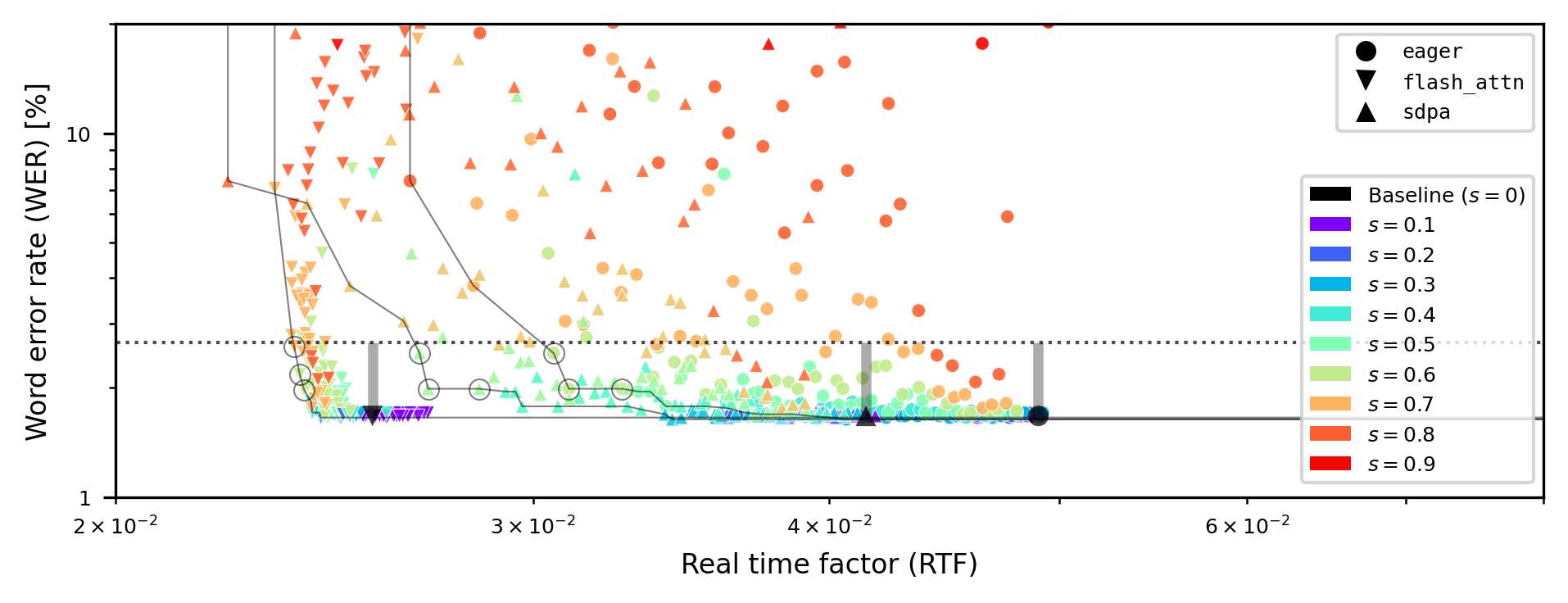}}
\caption{
    Empirical results of architecture search for the OpenAI \texttt{whisper-large-v3-turbo} model with different attention implementations. 
    Convention same as in Figure~\ref{fig:opt_res}.  
}
    \label{fig:pareto_attn}
\end{center}
\vskip -0.2in
\end{figure*}

To demonstrate the compatibility with kernel optimizations, we applied our encoder sparsification strategy to alternative attention implementations, including FlashAttention-2 and PyTorch SDPA. We plot the trade off between WER and RTF of \texttt{whisper-large-v3-turbo} using different attention implementations in Fig. \ref{fig:pareto_attn}. Similar to the eager attention implementation, the Pareto front of models with alternative attention implementation is also dominated by sparsities between 0.4 to 0.6, providing speedups within the accuracy degradation range. Because our algorithm needs raw attention scores from only one encoder layer, we compute that single layer with the standard eager attention and run every other layer with the accelerated FlashAttention-2 or SDPA kernels. Consequently the method inherits the full speedup benefits of the optimized kernels, confirming that our method and kernel-level acceleration are complementary.
\label{app:attn}

\begin{table*}[t!] 
\centering
\scriptsize

\begin{tabular}{l|l|l|l|l|l|l}
    \toprule
        Model & Sparsity & $\mathtt{mean}$ (ours) & $\mathtt{max}$ & $\mathtt{min}$ & $\mathtt{geometric\_mean}$&$\mathtt{random}$\\
        \hline \hline
        \multirow{8}{*}{\texttt{whisper-tiny} (6.266)} & 0.1 & \textbf{6.232} & 6.876 & 6.806 & 6.649 & 6.841\\
        & 0.2 & \textbf{6.127} & 7.990 & 16.884 & 6.719 & 6.823\\
        & 0.3 & \textbf{6.057} & 8.947 & 57.163 & 7.084 & 7.520\\
        & 0.5 & \textbf{6.144} & 10.496 & 202.576 & 8.285 & 15.352\\
        & 0.6 & \textbf{6.214} & 17.668 & 330.078 & 17.493 & 45.274\\
        & 0.7 & \textbf{6.632} & 14.117 & 445.152 & 49.939 & 61.741\\
        & 0.8 & \textbf{25.692} & 61.706 & 730.496 & 228.285 & 244.682\\
        & 0.9 & \textbf{159.077 }& 178.938 & 821.480 & 600.400 & 584.874\\
        \hline
        \multirow{8}{*}{\texttt{whisper-base} (4.352)} & 0.1 & \textbf{4.282} & 4.421 & 4.334 & 4.595 & 4.508\\
        & 0.2 & \textbf{4.230} & 4.299 & 5.048 & 4.752 & 4.648\\
        & 0.3 & \textbf{4.247} & 4.386 & 10.270 & 4.769 & 5.292\\
        & 0.5 & \textbf{4.247} & 4.769 & 61.793 & 5.292 & 12.533\\
        & 0.6 & \textbf{4.386} & 5.048 & 187.485 & 6.179 & 35.161\\
        & 0.7 & \textbf{4.665} & 6.719 & 348.947 & 43.725 & 114.221\\
        & 0.8 & \textbf{11.802} & 21.166 & 760.644 & 188.442 & 344.804\\
        & 0.9 & 86.510 & \textbf{75.857} & 1345.309 & 687.398 & 861.897\\
        \hline
        \multirow{8}{*}{\texttt{whisper-small} (3.064)} & 0.1 & 3.064 & \textbf{3.011} & 3.708 & 3.064 & 3.725\\
        & 0.2 & \textbf{3.046} & 3.081 & 3.272 & 3.081 & 3.220\\
        & 0.3 & \textbf{2.994} & 3.064 & 3.516 & 3.655 & 3.464\\
        & 0.5 & \textbf{2.924 }& 4.961 & 28.216 & 2.994 & 6.005\\
        & 0.6 & \textbf{2.959} & 15.213 & 102.402 & 7.502 & 21.410\\
        & 0.7 & 10.026 & 49.800 & 294.552 & \textbf{5.326 }& 138.120\\
        & 0.8 & \textbf{26.075 }& 261.410 & 596.014 & 78.016 & 347.050\\
        & 0.9 & \textbf{229.017 }& 1155.039 & 1052.916 & 556.588 & 648.721\\
        \hline
        \multirow{8}{*}{\texttt{whisper-medium} (2.019)} & 0.1 &\textbf{ 1.967} & 2.106 & 2.054 & \textbf{1.967} & 2.019\\
        & 0.2 & \textbf{2.002} & 2.089 & 2.141 & \textbf{2.002} & 2.124\\
        & 0.3 & \textbf{1.984} & 2.193 & 15.161 & \textbf{1.984} & 2.228\\
        & 0.5 & \textbf{2.002} & 10.357 & 127.746 & \textbf{2.002} & 3.307\\
        & 0.6 & \textbf{2.071} & 31.158 & 279.582 & 2.263 & 5.379\\
        & 0.7 & \textbf{2.332 }& 51.506 & 510.218 & 3.394 & 28.077\\
        & 0.8 & \textbf{39.130 }& 120.035 & 835.892 & 56.623 & 169.974\\
        & 0.9 & 384.421 & \textbf{95.352} & 613.020 & 457.685 & 610.653\\
        \hline
        \multirow{8}{*}{\texttt{whisper-large-v3-turbo} (1.671)} & 0.1 & \textbf{1.671} & 1.741 & 1.793 & \textbf{1.671} & 1.845\\
        & 0.2 & \textbf{1.671} & 1.775 & 1.932 &\textbf{ 1.671 }& 2.002\\
        & 0.3 & \textbf{1.688} & 1.793 & 2.419 & \textbf{1.688} & 2.350\\
        & 0.5 & 1.775 & 8.651 & 14.151 & \textbf{1.723} & 13.908\\
        & 0.6 & 1.810 & 24.352 & 73.299 & \textbf{1.706} & 45.379\\
        & 0.7 & \textbf{1.880} & 102.367 & 80.470 & 1.915 & 185.396\\
        & 0.8 & 5.257 & 76.258 & 195.753 &\textbf{ 2.507} & 433.542\\
        & 0.9 & 75.144 & 238.416 & 327.850 & \textbf{49.678} & 921.410\\
        \hline
        \multirow{8}{*}{\texttt{whisper-large-v3} (1.514)} & 0.1 & \textbf{1.497} & 7.467 & 1.775 & \textbf{1.497} & 1.584\\
        & 0.2 & \textbf{1.497} & 33.020 & 2.489 & \textbf{1.497} & 1.706\\
        & 0.3 & \textbf{1.514} & 83.046 & 4.595 & \textbf{1.514} & 1.897\\
        & 0.5 & 1.549 & 96.240 & 56.971 & \textbf{1.532} & 3.220\\
        & 0.6 & \textbf{1.584} & 96.101 & 161.706 & 1.810 & 9.434\\
        & 0.7 & 8.634 & 96.501 & 213.281 & \textbf{6.649} & 47.032\\
        & 0.8 & \textbf{4.352} &  & 176.936 & 18.799 & 214.900\\
        & 0.9 & 113.768 &  & 95.735 & \textbf{85.953} & 394.030\\

    \bottomrule
\end{tabular}
\caption{WER (\%) for models at different sparsities using various aggregation functions. Entries in bold presents the best results with lowest WER. Number in parenthesis next to the model name is the baseline WER. WER at $\text{sparsity}\geq0.8$ for \texttt{whisper-large-v3} using $max$ aggregation function are missing as they are trapped in infinite generation loops. }
\label{table:agg_func}
\end{table*}


\end{document}